\documentclass[letter,10pt,conference]{ieeeconf} 

\IEEEoverridecommandlockouts 
\usepackage{graphics} 
\usepackage{epsfig} 
\usepackage{mathptmx} 
\usepackage{times}
\usepackage{amsmath}
\usepackage{amssymb} 
\usepackage{comment}
\usepackage{xcolor}
\usepackage{makecell}
\usepackage{subcaption}
\usepackage{tikz}
\usepackage{color}
\usepackage[export]{adjustbox} 

\DeclareMathOperator*{\argmax}{argmax}

\title{\LARGE \bf Quality-Adaptive Multi-UAV 3D Reconstruction \\ with Sparse Workload Redistribution}

\author{Benjamin Sportich¹, Kenza Boubakri¹, Olivier Simonin¹, Alessandro Renzaglia¹
\thanks{$^{1}$Inria, INSA Lyon, CITI, 69621 Villeurbanne, France. Email: {\tt\small fistname.lastname@inria.fr}}
\thanks{This work has been funded by the Agence Nationale de la Recherche (ANR), AVENUE project, grant ANR-22-CE33-0004.}
}

\begin{document}

\maketitle
\thispagestyle{empty}
\pagestyle{empty}

\begin{abstract}
3D reconstruction of unknown environments is a key application in robotics but is severely limited by the computational and energy capabilities of current aerial platforms. Deploying multiple UAVs and providing efficient and scalable path planning strategies are common approaches, but effective online coordination among UAVs remains a significant challenge. To address this problem, we propose a quality-adaptive decentralized decision-making strategy to build a 3D map with user-defined degrees of fidelity. The approach integrates a quality-oriented criterion based on TSDF confidence into view generation and information gain estimation to produce viewpoints consistent with the desired fidelity target. Additionally, we employ two levels of coordination: a penalty factor in the viewpoint evaluation to encourage local dispersion among the UAVs and a global imbalance correction mechanism. The latter, based on regularized clustering and optimal task assignment, is only triggered when an unbalanced configuration relative to high-information regions is detected.
Simulation results demonstrate that the proposed method improves path efficiency compared to state-of-the-art multi-UAV exploration approaches, while also achieving higher-fidelity reconstructions in terms of coverage and accuracy.
We make our code publicly available to the community\footnote[2]{https://github.com/Chroma-CITI/QUARTZ}.

\end{abstract}

\section{Introduction}
Autonomous Unmanned Aerial Vehicles (UAVs) are increasingly used to provide high-fidelity 3D models of complex and initially unknown environments for applications such as inspection, monitoring, and search-and-rescue \cite{peng2019}, \cite{hardouin2023}. Producing 3D reconstructions with multiple UAVs requires balancing two often-conflicting objectives: cover the scene quickly to minimize mission time and energy consumption, and achieve quality requirements via viewpoints associated with accurate and complete surface estimates. Both single and multi-robot reconstruction planning for exploration have made strong progress, but they typically optimize one objective at the expense of the other: exploration methods prioritize coverage and travel efficiency at the cost of map precision, while reconstruction methods prioritize viewpoint quality and geometric accuracy at the cost of mission time. Additionally, at the fleet level, these strategies also need to be translated into efficient coordination solutions to deal with redundant observations and unbalanced workloads, while keeping computational costs and communication burdens at acceptable levels. 

In this paper, we tackle these two objectives simultaneously by proposing a novel quality-adapted Next-Best-View (NBV) formulation for a fleet of cooperative UAVs. Our approach treats reconstruction fidelity as an explicit, tunable objective derived from voxel confidence in a Truncated Signed Distance Field (TSDF) map, which allows the user to weigh speed and accuracy and even specify spatially varying fidelity requirements. 
At the decision level, we combine a decentralized NBV selector, driven by information gain, navigation, and coordination terms, with a sparse, fleet-aware workload redistribution mechanism that corrects spatial imbalances only when necessary. This results in an efficient coordinated exploration process where additional observations are taken whenever the confidence levels on the online built map do not reach predefined objectives to ensure a complete and quality-aware final surface reconstruction.

The main contributions can be summarized as follows:

\begin{itemize}
    \item A quality-adaptive next-best view planning problem formulation, linking exploration and reconstruction through a continuous control parameter based on TSDF confidence. In this model, region-specific priorities within the same map can also be defined. 
    \item A decentralized fleet-aware decision-making framework requiring no prior or preliminary survey, with a sparse dynamic workload redistribution mechanism triggered by spatial imbalance detection. 
    \item An extensive performance evaluation demonstrating improved path efficiency against competitive exploration methods while maintaining high reconstruction fidelity.  
\end{itemize}

\section{Related Work}
\label{sec:rel_work}

Optimizing coverage for exploration is often insufficient to obtain a high-confidence surface reconstruction, which benefits from diverse viewing angles aligned with the surface normals at appropriate viewing distances \cite{peng2019}, \cite{Schmid2020}. However, exploration strategies also benefit from incorporating quality constraints, as they can improve path efficiency or prevent repeated traversal, as done in \cite{epic}, which constrains observation angles and discriminates between sufficiently and poorly observed areas. Reconstruction strategies, despite their focus on accuracy, are bound by the same computational and energy limits as exploration methods. Thus, they also have to minimize flight time, even if they cannot traverse the environment as aggressively.
These overlapping constraints have led the field to adopt similar tools for viewpoint generation \cite{Schmid2020}, path selection \cite{fuel}, and environment representation \cite{fiesta}, \cite{voxblox} but most approaches predominantly optimize either the coverage or the reconstruction accuracy objective. In our previous work \cite{Sportich2026ICUAS}, we proposed a first solution for a single UAV that simultaneously takes into account both criteria.

Within this context, numerous exploration and reconstruction solutions that exploit multi-robot cooperation have been proposed, but efficient coordination to minimize redundancy remains a challenge. Some recent exploration strategies rely on topological maps to distribute the workload. For instance, \cite{racer} divides the space using \textit{hgrid} cells that are allocated through a VRP calculation to balance the paths locally between robot pairs without fleet-wide optimization. \cite{distributed_vor} creates map subdivisions through Voronoi graph partitioning to avoid UAVs covering the same area, but without control over travel distance balance. In \cite{fast_forests}, UAVs can dynamically switch between \textit{explore} and \textit{coverage} modes depending on their surroundings. Areas of attraction are assigned, but again only between robot pairs through dedicated communication. 

Some reconstruction methods tend to employ the \textit{explore then exploit} strategy, which consists of first obtaining a coarse map of the environment before generating globally optimized viewpoints that can be traversed all at once. Already popular in single-robot applications \cite{peng2019}, it was adapted to multi-robot deployment in \cite{sweep_your_map} and \cite{soar}. The former generates and distributes large sweeps around poorly covered areas using a VRP model after a first single-drone high-altitude lawnmower pattern provides a rough map. The latter requires a heterogeneous fleet with specific sensor capabilities for task decomposition: a LiDAR-equipped UAV dubbed the \textit{explorer} UAV and camera-equipped UAVs named \textit{photographers}. The explorer is launched first, and incremental viewpoints are then distributed to the photographers via a Consistent-MDTSP method. Both allow for efficient task repartition but require a central server to compute the plan distribution. \cite{hardouin2023} proceeds in one sweep but uses a VRP model on a partially known map that has to be recomputed multiple times, potentially leading to high computational costs. In contrast, our decentralized decision-making accounts for the whole team through a dispersion strategy while spatially rebalancing the UAVs when necessary.

\section{Problem Formulation}
\label{sec:problem}
We consider a team of $N$ cooperative aerial robots equipped with a depth sensor with a limited Field-of-View (FoV) characterized by a maximum and minimum range $r_{max}$ and $r_{min}$, respectively. At all times, we assume access to a perfect localization system in a common reference frame and perfect communication. The robots' goal is to reconstruct the 3D surface of an unknown environment in the minimum time possible while achieving predefined quality objectives. 
We call $Z$ the quality measure of the 3D reconstruction, based on the built map $M$, which will be later formally defined.
Let $q^r \in SE(3)$ be the pose of the robot $r$, with $r \in [1,N]$, and $q^r_{0}$ its initial configuration. $q^r$ is defined by a position $s^r_q \, \in \, \mathbb{R}^{3}$ and orientation $\phi^r_q \, \in \, \mathbb{R}^{3}$. 

The map $M$ is a volumetric map of the environment based on a discretized voxel grid of the 3D space. A voxel, of size $d_s$, can refer to \textit{free}, \textit{occupied}, or \textit{unknown} space. We follow the general definition of surface frontiers given in \cite{hardouin2023}:

\textbf{Definition 1 (Incomplete Surface Element):}
An
\textit{Incomplete Surface Element} (\textit{ISE}), is a voxel $i \in M$ lying on the surface at a frontier, near both the unknown and empty space. 
More formally, a voxel $i \in ISE$ if and only if:
\begin{itemize}
    \item $i$ is \textit{empty}, 
    \item $\exists \, i' \in \mathcal{N}^6_i \; \text{s.t. $i'$ is \textit{unknown},}$
    \item $\exists \; i'' \in \mathcal{N}^{18}_i \; \text{s.t. $i''$ is \textit{occupied},}$
\end{itemize}
where $\mathcal{N}^6_i$ and $\mathcal{N}^{18}_i$ denote the 6- and 18-connected voxel neighborhoods of $i$, respectively. The definition of an empty, unknown, occupied voxel is dependent on the map representation and we will detail this further below.

\textbf{Definition 2 (Remaining Incomplete Surface):} Let $Q$ be the set of all collision-free configurations of an agent, and let $Q_{ISE} \subseteq Q$ be the set of all configurations from which an ISE $i \in ISE$ can be scanned. The remaining incomplete surface is then defined as:
\begin{equation}
ISE_{\text{rem}} = \bigcup_{i \in ISE} \{ i \mid Q_{ISE}= \emptyset \}.
\end{equation}

\subsection{Quality of the reconstruction}

To represent the voxel maps, we choose to use Truncated Signed Distance Fields (TSDF), where each voxel $i \in M$ is represented by an aggregated projected distance $d_i$ to the closest obstacle, and a weight $w_i$ acting as a measure of confidence of this distance.

Within a TSDF representation, a voxel $i$ is considered:
\begin{itemize}
    \item empty, if $w_i > 0$ and $d_i > d_s$ (voxel size),
    \item occupied, if  $w_i > 0$ and $d_i < d_s$,
    \item unknown, if $w_i = 0$.
\end{itemize} 

The seminal works \cite{kinect} and \cite{nguyen2012modeling} have extensively studied the impact of sensor uncertainty on reconstruction problems and how to incorporate it into the weight update phase to reflect map confidence. These results are now commonly used by state-of-the-art TSDF libraries, such as \cite{voxblox} and \cite{pan2022_voxfield}. Therefore, although TSDF weights are not a direct geometric error metric, they can be considered a reliable proxy for evaluating the accuracy of the built map. 
In this work, the weights of the TSDF representation, $w_i$, are exploited to define a function $Z(M)$: 
\begin{equation}
Z(M) = \xi (M) \, \bar{Z} = \xi (M) \sum_{i \in M_S }{ \frac{w_i}{ |M_S| }} 
\label{quality_function}
\end{equation}
where $M_S$ is the set of observed surface voxels in $M$ and $\xi (M) = 0 $ if $\exists \, i$ such that $i \in ISE$ and $i \notin ISE_{\text{rem}}$, and $ \xi (M) = 1 $ in all other cases. As a result, $ \xi (M) $ evaluates the completeness of the surface reconstruction and $\bar{Z}$ represents the average confidence in the obtained map.
It is worth noticing that the TSDF representation has been proved to be equivalent to the least squares minimization of squared distances between points on the range surfaces and points on the desired reconstruction under the specific conditions of an orthographic sensor and a range error independently distributed along the line of sight of the sensors  \cite{tsdf_foundation}. Hence, our function $Z$ can be considered as an efficient proxy for a quality measure of the reconstructed isosurface.

Using the quality function in eq.~\eqref{quality_function} can however lead to overestimating the overall accuracy of a map, where some parts of unnecessarily high confidence compensate for lower-accuracy areas. We thus define the following objective function $Z_{sat}(M;w^*)$:
\begin{equation}
Z_{sat}(M;w^*) = \xi (M) \, \bar{Z}_{sat} = \xi (M) \sum_{i \in M_S}{ \frac{\text{min(}w_i,w^* )}{ |M_S| }} 
\label{quality_function_2}
\end{equation}
where $w^{*}$ is a predefined confidence goal. 
Using this formulation, we can formulate our problem as follows.

\textbf{Multi-Robot Quality-Aware Mapping (MR-QAM) problem:} Considering a team of $N$ UAVs with initial configurations $q^r_0 \in Q, r \in \{1, 2,  \ldots, N\} $, the MR-QAM asks to identify $S = (  \hat{Q} , \hat{P}_{\hat{Q}} )$ a set of poses and the corresponding collision-free paths which allow the robots to reach a quality of reconstruction $ Z^*:= \text{max}(Z_{sat})$ on the fused map in a minimum time. 

Note that, based on eq.~\eqref{quality_function_2}, $Z^*:=\text{max}(Z_{sat})=w^*$. In this formulation, fast exploration and high-quality reconstruction can be interpreted as special cases of the MR-QAM problem where $Z^*$ tends to $0$ and $1$, respectively.
Following this formalism, we then propose a new definition to identify surface voxels that do not meet the confidence objective.

\textbf{Definition 3 (Low Quality Surface Element):} an \textit{LQSE}, is a voxel $i \in M$ lying on the surface such as $w_i<w^*$.

Based on Definitions 2 and 3, we finally define the set of \textbf{Active Surface Elements} $ASE$ in $M$ as the union of $LQSE$s and $ISE$s.
This formulation also allows easily checking whether the completeness and confidence requirements have been met because:
\begin{equation}
    Z_{sat}(M;w^*)=Z^* \; \iff \; ASE = \emptyset \,.
\end{equation}
The corresponding minimum threshold $Z^* = w^*$ can be tailored depending on the application and can even be set to different values in different predefined subregions of the map. Although the MR-QAM optimization problem cannot be solved directly, the following section presents an online NBV planning approach that aims to achieve the same objective.

\section{Proposed Methodology} \label{sec:approach}
Our approach follows the NBV framework of iteratively creating informed viewpoints based on a partial volumetric map and then selecting the best configuration under a varied set of constraints. However, the approach we present in this paper is twofold. First, we explicitly incorporate a user-defined quality objective in the determination of locally optimal viewpoints. Additionally, we use a fleet-aware decentralized decision-making strategy that operates on two complementary levels. The first layer encourages spatial separation of the UAVs in the environment using a repulsion term based on the overlap of influence spheres. However, simple repulsive terms can still lead to poor dispersion depending on the initial positions and the topology of the environment. To prevent this, it is coupled with a global workload correction step when an imbalance or an unfavourable spatial configuration of the UAVs in the environment is detected. Triggered by a fleet member, this process is implemented via a combination of balanced clustering of active surface elements and MRTA assignment.

\begin{figure}[tb]
    \centering
    \includegraphics[width=.9\columnwidth]{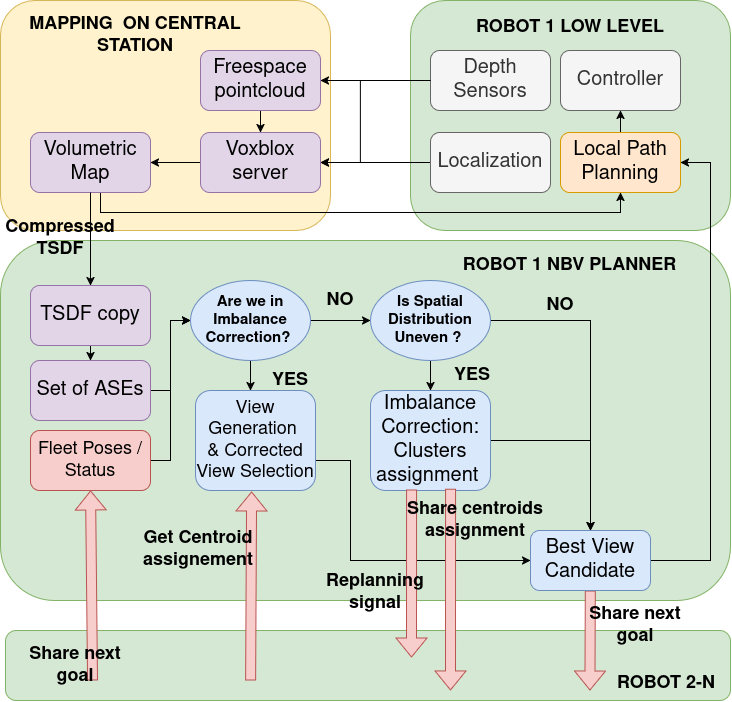}
    \caption{Proposed pipeline architecture. \vspace{-0.2cm}}
    \label{fig:software_stack}
\end{figure}

An overview of our architecture is shown in Figure~\ref{fig:software_stack}. 
The point clouds perceived by the onboard sensor are integrated with the Voxblox library \cite{voxblox} to build a TSDF and an ESDF simultaneously. The map fusion is performed on a base station and aggregating the UAVs' individual point clouds. At regular intervals, the fused map is transmitted back to the UAVs using the Voxblox compressed SDF representation. The NBV planner system then uses these representations to determine informative viewpoints. The low-level planner generates collision-free trajectories accordingly. 
Each UAV broadcasts its current pose, its committed next target pose, and a binary flag indicating availability for fleet-level replanning. The current pose is used for low-level collision avoidance, whereas the target and binary flag enable the decentralized decision process, within the dispersion component and for UAV synchronization, respectively.

\subsection{Quality-oriented Viewpoint Generation}
The first step of our approach is to identify an aimed observation distance from the surface to reconstruct, which will then be used to generate and evaluate viewpoint candidates.
Our objective function $Z_{sat}$, defined in eq.~\eqref{quality_function_2}, is built on the TSDF representation and exploits the associated confidence estimates encoded in the weights $w_i$. Leveraging the results presented in \cite{nguyen2012modeling}, the confidence of a given measurement is typically set to $1/d^2$, where $d$ is the measurement depth. For example, this is the case for the TSDF library Voxblox \cite{voxblox} that we adopt in our work.
Based on this model, to obtain a confidence $w^*$ for a given surface voxel would be sufficient to observe it from a distance $d^*$ equal to:
\begin{equation}
    d^{*} = \min{ (\sqrt{ 1 / w^{*}}, r_{max} )} \,.
\end{equation}
On the map level, this relation implies that, if all surface voxels are observed at a distance of $d^{*}$ or less at least once, then $Z \geq Z^{*}$.
When $Z^{*}$ is close to 0, $d^{*}$ would correspond to the maximum range of the sensor, meaning voxels can be observed from any distance and still satisfy the quality requirements. On the other hand, if $Z^{*}$ is close to 1, then $d^{*}$ will be close to the sensor's minimum range. It is important to note that a given confidence level could also be achieved with multiple observations obtained from distances greater than $d^{*}$ (see \cite{voxblox} for more details on the weight update rule).
Based on these considerations, we use $d^*$ as our reference observation distance to guide both view generation and evaluation. In practice, we define a viewing distance interval $[d^*-\eta ; d^*+\eta]$ with $\eta>0$, and our strategy aims for a complete exploration via the generation and selection of views within this viewing distance interval.


When the robot starts its exploration or is about to reach the current goal position and orientation, the UAV examines its map, and all the $ASE$s voxels are identified. A view candidate $v_e \in SE(3)$, defined by a position $s_v \, \in \, \mathbb{R}^{3}$ and an orientation $\phi_v \, \in \, \mathbb{R}^{3}$, is generated for each of these voxels $e \in ASE$. 
Three points are sampled along the normal of the active surface element\footnote{Normal vectors are obtained via the method presented in \cite{hardouin2023}, which exploits all 26 neighboring voxels and incorporates uncertainty.} at different viewing distances $d$: in a close range $d <  d^*-\eta$, in an optimal range ($d \in [d^*-\eta ; d^*+\eta] $), and at long range ($d >  d^*+\eta$). Points around the optimal distance, then close-range, and long-range are checked for safety in this order, which is in accordance with the satisfaction of the aimed quality threshold. Once a valid position is found, the process is stopped and moves to the next element. The final set of generated candidate viewpoints for the UAV $r$ is called $V_r$. 
The view generation process can only start once the $ASE$ spatial distribution is balanced (see subsection \ref{sec:approach}-C for more details).

\subsection{First Coordination Layer: View Evaluation Function}

Each UAV evaluates the view candidates generated in $V_r$ according to an information gain component $J_I$, a navigation component $J_N$, and a coordination component $J_C$. The selected view $v^*$ for the robot $r$ is the view that maximizes the function $J_r$ that combines these three terms, i.e.: 
\begin{equation}
    v^* = \argmax_{v \in V_r} \; J_r(v) = \argmax_{v \in V_r} \;\alpha \, J_I(v) + \beta \, J_N(v) + \gamma \, J_C(r,v)
    \label{layer_function}
\end{equation}
Once $v^*$ is identified, the UAV plans a trajectory using the low-level path planner and navigates towards that pose. This process repeats iteratively and the UAV moves from a locally optimal viewpoint to the next until no new views are identified.  All $J_I$, $J_N$ and $J_C$ have normalized values in $[0,1]$.
We set $\gamma$ as the dominant weight to prioritize minimizing redundancy over individual coverage performance.
Considering that the size of $ASE$ can grow exponentially with the size of the map, our evaluation function
was designed to both discriminate efficiently and be computationally tractable.

\subsubsection{Information Gain Component}
Our information gain component is guided by the objective quality $Z^*$ and considers viewing distance and visibility probabilities due to occlusions to differentiate views.
A voxel $i$ is generally called \textit{occluded} from a view candidate $v$ if along the ray from $v$ to $i$ there is at least one occupied voxel. 
We define $ASE_{v} \subset ASE$ the set of non-occluded active surface elements from the pose of the view candidate $v$, and $\delta_{v,e}$ a binary variable that equals 1 if $e \in ASE_{v}$ and 0 otherwise. However, this definition does not take into account the possible occlusion of unknown voxels along the ray that \textit{could} be occupied. 
We thus define the function $vis(v,e)$ estimating the visibility of the target surface element $e$ from the view candidate $v$, considering the potential occlusions due to unknown voxels, as follows: 
\begin{equation}
    vis(v,e) =  \delta_{v,e}\, exp({-u})
\end{equation}
with $u$ being the number of unknown voxels along the ray from the sensor pose $v$ to the target surface element $e$. We thus define the information gain function $J_I(v)$ as:
\begin{equation}
    J_I(v) = \frac{1}{ \max\limits_{v \in V_r} |ASE_v|} \sum_{e \in ASE} h_v(e) \, vis(v,e) 
\end{equation}
where 
\begin{equation} 
h_v(e) = \begin{cases}
1 & \text{if } |d(e,s_v)  - d^{*}| < \eta  \\
0.5 & \text{if }  d(e,s_v) < d^* - \eta  \\
0.5 \left(1-\frac{d(e,s_v)}{r_{max}}\right) & \text{if }   d(e,s_v) > d^* + \eta
\end{cases}
\end{equation}
The function $h$ prioritizes view candidates that observe active surface elements from a distance close to the preferred one $d^*$. Views at farther distances are penalized, as they do not satisfy the quality criteria. While closer viewing distances do, they have fewer voxels in the field of view and are therefore also penalized, though less significantly.

\subsubsection{Navigation component}

During the exploration, a major fallback is to return to previously visited areas to observe a missed portion of the surface, generating additional and unnecessary travel. Furthermore, the position $s_v$ of the view candidate $v$ can be close in distance to the actual robot position $s_q \in \mathbb{R}^{3}$ but facing the opposite direction. Moving to that view would incur a costly change in direction with a trajectory that cannot be a straight line. To generate efficient trajectories, we include both of these aspects in the design of our navigation component $J_N$:
\begin{equation}
    J_N (q,v) = \psi(q,v) \frac{\min\limits_{v \in V_r} || s_v - s_q || }{|| s_v - s_q ||}
\end{equation}
where \begin{equation}
      \psi(q,v) = 1 - \frac{1}{\pi} \arccos{\left( \frac{ { vel }^T }{ || vel || } \cdot \frac{  s_v - s_q  }{  || s_v - s_q ||  }\right) }
\end{equation}
with $vel$ being the velocity vector of the robot.

\subsubsection{Coordination component}

The coordination component is designed as a repulsion force aiming to disperse the robot in the 3D space, allowing for an efficient task division, diminishing redundancy and collision probabilities. The dispersion function, for each robot, penalizes views that are too close to other robots' next goal, in accordance with each robot's quality-informed viewing distance range. Thus, selected views outside of the other robots' sensors' reach are prioritized, effectively preventing spatial overlap.  

We use a formulation based on the intersection of two spheres centered in the robots' present or future position: 
\begin{equation}
    J_C(r,v) = 1 - \frac{1}{|N-1|}\sum_{i =1 , i \neq r}^{N}\frac{| \textit{Sphere}( v,d^*) \cap  \textit{Sphere}(g_i,d^*)| }{|\textit{Sphere}(s^r , d^*)|}
\end{equation}
where $\textit{Sphere(a,b)}$ is a sphere centered in $a$ and of radius $b$, $d^*$ is the optimal viewing distance for the robot $r$, 
$g_i$ is the next goal of robot $i$, and $s^r$ the current position of robot $r$. 

\subsection{Second Coordination Layer: Imbalance Correction}

\begin{figure*}[htbp]
    \centering

    \begin{minipage}[t]{0.4\textwidth}
        \centering
        \includegraphics[width=\linewidth]{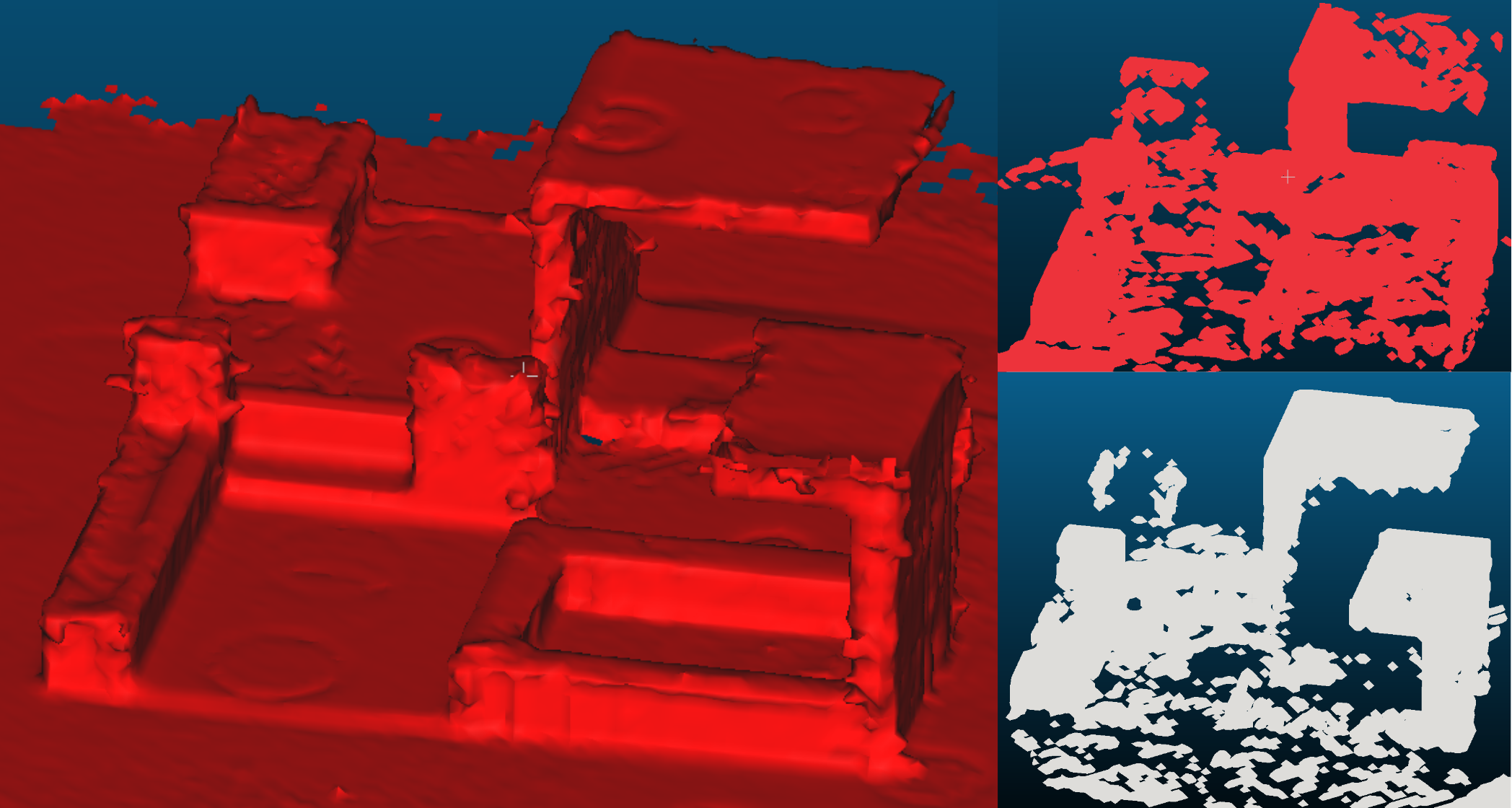}
        \captionof{figure}{Final mesh of "Structure" obtained with our method (left), RACER \cite{racer} (right-up), and FAME \cite{fast_forests} (right-down).}
        \label{fig:qualitative_mesh}
    \end{minipage}
    \hfill
    \begin{minipage}[t]{0.56\textwidth}
        \centering
        \includegraphics[width=\linewidth]{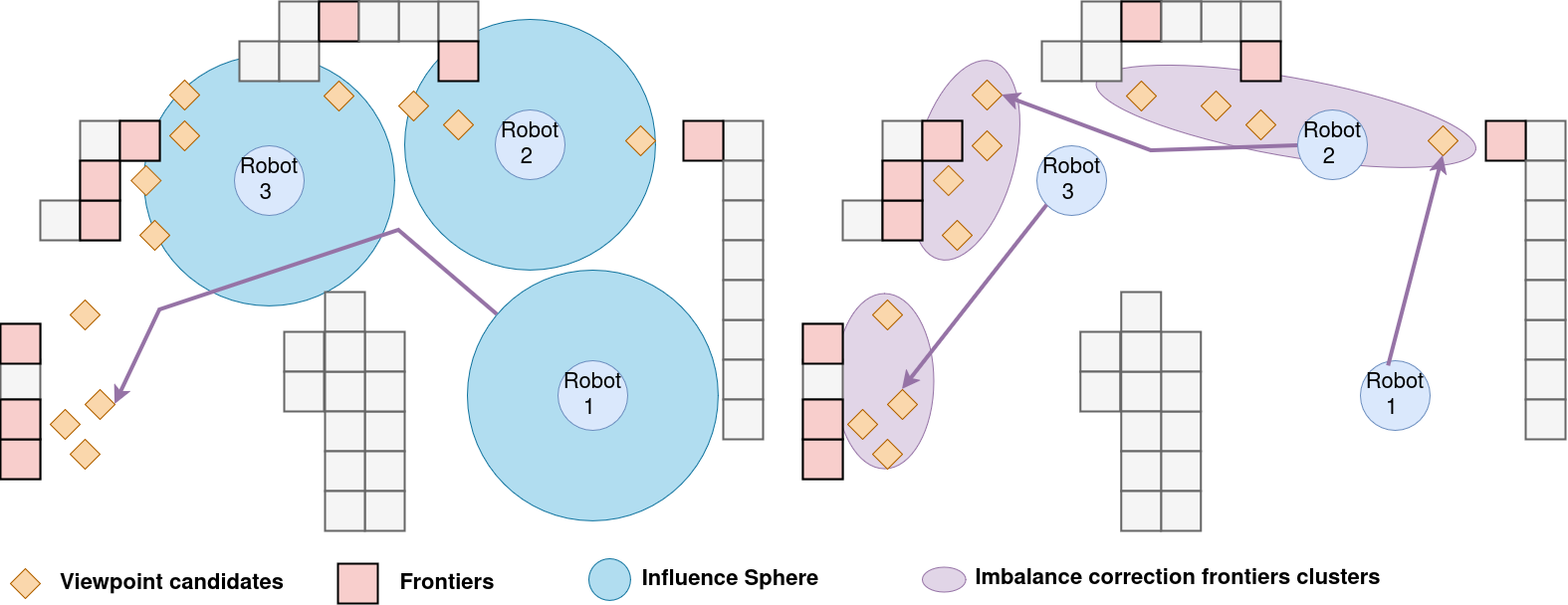}
        \captionof{figure}{Decision process for viewpoint selection without imbalance correction (left) and with imbalance correction (right).}
        \label{fig:imbalance_correction}
    \end{minipage}

\end{figure*}

Despite being widespread and proven strategies, dispersion functions for multi-robot systems are prone to get stuck in local minima in environments with complex geometry or when one robot reaches a dead end during the exploration.
To prevent this common fallback, we implemented a reconfiguration process that triggers when the spatial distribution of active surface elements among robots appears too uneven (see Fig.~\ref{fig:imbalance_correction}). These elements are divided into approximately equal sets and each robot is then assigned a corresponding subregion of the map to select its next viewpoints from, recalibrating the robot team in a more balanced configuration. Our core assumption is that a uniform distribution of active surface elements across the robot fleet during the mission allows for greater adaptability in unknown environments than standard strategies, while also improving exploration speed by balancing the workload. If the spatial distribution, which is computed right after the detection of active surface elements, is judged uneven, the UAV will trigger the correction. Otherwise, it will proceed to generating and selecting views using eq.~\eqref{layer_function}.

\subsubsection{Imbalance Evaluation} The robot $r$ detects a spatial imbalance by computing the following function $f_r$: 
\begin{equation}
\label{eval_eq}
    f_r(x_1,...,x_{N}) = \frac{1}{2} \sum_{i= 1 }^{N} \left(\frac{|ASE|}{N} - | \underset{ASE}{Vor} (x_i)| \right)^{2}
\end{equation}
where $x_i = g_i$ if $i \neq r$ and $ x_r = s_q^r$, and $\underset{ASE}{Vor}(x_i)$ is the set of $ASE$s in the Voronoi region generated by $x_i,\ i \in [1,N]$, i.e.:
\begin{equation}
    { e \in \underset{ASE}{Vor}(x_i) \iff \lVert e - x_i \rVert \leq \lVert e - x_j \rVert \, \forall j \in [1,N] \setminus \{i\} }\,.
    \label{voronoi_eq}
\end{equation} 
If the current value of $f_r$ is higher than a certain threshold $f_{min}$, the reconfiguration process is launched. 
It consists of three different steps: a balanced clustering of the elements of the $ASE$ set, the assignment of a part of the map, and the corrected selection.  
The balanced clustering and the sub-map assignment are performed by the robot detecting an unequal distribution of $ASE$s, whereas all the robots are involved in the solution deployment, which consists of selecting views in their assigned area for a predetermined number of steps.

To identify a suitable threshold $f_{min}$, we consider the scenario where one single UAV has a deviation of $\Delta$ compared to the optimal allocation $|ASE|/N$. In this case, and using eq.~\eqref{eval_eq}, it can be shown that
\begin{equation}
    f_r(x_i;\Delta) \geq \bar{f_r}(\Delta)=\frac{\Delta^{2}N}{2(N-1)} \,.
\end{equation}
Defining $f_{min}=\bar{f_r}(\Delta)$ makes it easier to set the threshold since it has a more physically interpretable meaning related to the deviation $\Delta$. In particular, in our solution, we set $\Delta = \frac{|ASE|}{N(N+1)}$, meaning any configuration with an UAV that has allocated $\frac{|ASE|}{N+1}$ or worse, would trigger the replanning.

\subsubsection{Balanced Clustering}
To cluster the active surface elements, we use the regularized k-means algorithm \cite{regularized_kmeans}, a multi-objective variant of a standard k-means that adds a secondary constraint to force the creation of same-size clusters from the $ASE$ set. We create a number of clusters equal to the number of robots $N$ in the team.

\subsubsection{Cluster Assignment} Instead of assigning the clusters themselves, we assign their centroids that encode all the necessary spatial information. We note the centroid assigned to the robot $r$  $\text{Centroid}(r)$.
The assignment is done using the optimal Kuhn–Munkres algorithm on a cost matrix filled with the distances between $x_i, i \in [1,N]$ and the centroids. These distances are approximated using a path searching algorithm on a topological grid map of cell sizes equal to 4 times the voxel size, updated at the same time as the $ASE$.  
Once performed, the UAV can transmit the assignment to the fleet using a compact message of $N$ 3D points. 

\subsubsection{Corrected Selection} After the centroid assignment phase, each robot $r$ can resume its view generation and selection process. However, the computation of $J_C$ from \eqref{layer_function} is changed for $k$ iterations and we have for the robot $r$:
\begin{equation}
    J_{r,C}(v_e) = 0 \quad \forall\,e \notin \underset{ASE}{Vor}( Centroid(r))  
\end{equation}
with $v_e$ the view candidate generated from the active surface element $e$. In practice, views generated from inside $Vor(Centroid(r))$ are advantaged while others are penalized.
This corrected selection process, repeated for $k$ iterations (see section \ref{sec:results} for more details), guides robots toward more balanced regions and prevents unbalanced workloads, which would lead to longer mission times. 
Once the $k$ iterations elapsed, the partition will refine naturally as $ASE$ evolves and the dispersion component takes over.

\subsubsection{Synchronization} To avoid planning conflicts, a synchronization process ensures that the correction scheme is not launched by multiple robots at the same time. Once a UAV detects an imbalance, it updates its status to \textit{replanning}, which is shared with the fleet. When the other receives that status, they can no longer launch an imbalance correction until the one launched by the first robot is resolved. 

\subsection{Implementing Multiple Quality Requirements}

It is important to note that our approach supports the implementation of multiple quality requirements. Different $Z^*_i$ can be indeed defined over distinct subregions $M_i \subset M$, where $M_i$ form a partition of $M$, i.e., $M_i \cap M_j = \emptyset$ for $i \neq j$, and $\bigcup_i M_i = M$. In this scenario, the system operates as previously described, with minor adaptations. During view generation, an active surface element generates a view candidate with a $d^*_i$ corresponding to the subregion $M_i$ in which it lies. Similarly, the information gain is computed voxel-wise: for each voxel contributing to the gain, the corresponding regional distance $d^*_i$ associated with the subregion $M_i$ to which the voxel belongs is used in the evaluation. Finally, the coordination term $J_C$ considers the appropriate $d^*_i$ in the computation of the volume of each sphere, depending on which subregion $M_i$ of the general map $M$ each sphere is centered in. As this can lead to overlapping spheres based in different subregions, the denominator is $Sphere(s^r, \max_i d^*_i)$.

\subsection{Local Planning}
To represent the 3D space as a 3D map, we use ETHZ ASL's Voxblox representation \cite{voxblox}, which builds a TSDF and an ESDF concomitantly. The TSDF is used for the reconstruction reasoning and the ESDF by the navigation component to determine the freespace. We use the local planner from \cite{oleynikova2018safe} to generate dynamically feasible trajectories and replan in real-time to avoid obstacles. The local planner has two main components: first, the local trajectory optimizer that computes the best path between two waypoints without collisions. If found, this trajectory is then smoothed out to respect the velocity and acceleration constraints \cite{oleynikova2018safe}. Then, the intermediate goal selector creates sub-sections of the path between the current pose of the UAV and the goal and sends them one by one to the UAV's controller.

To improve the UAV's navigation and obstacle avoidance, a freespace point cloud is computed negatively from the UAV sensor's original point cloud.
This helps to detect the navigable environment and is also used within Voxblox to enhance the quality of the ESDF and TSDF. 
Finally, a collision checker runs at 20 Hz to prevent collisions with other UAVs and obstacles.
To this end, we use a bounding box in the direction of travel, whose size adapts proportionally to the velocity vector for collision detection.

\section{Numerical Experiments}
\label{sec:results}

In this section, we test our method in different simulated environments with varying characteristics.
To assess the efficiency of our approach, we compare its performance against two state-of-the-art methods: RACER \cite{racer}, and FAME \cite{fast_forests}. Both are multi-robot exploration strategies implementing decentralized decision making. RACER and FAME were primarily designed for efficient exploration rather than reconstruction fidelity, but they represent strong decentralized baselines and are widely used to produce volumetric maps.

\subsection{Simulation Setup}
Performance evaluations were performed in ROS1-Gazebo simulations. All methods use a quadrotor UAV equipped with a 3D LIDAR providing a full 360-degree horizontal field of view (FOV) and a vertical FOV ranging from -35 to +30 degrees. The UAV model is based on an Intelaero UAV, using identical motor placement and dimensions. To approximate sensor imperfections, we add to depth measurements a zero-mean Gaussian noise with a standard deviation increasing quadratically with distance ($\approx$1.5\,cm at 10\,m).
We assume perfect localization as in all the major works of the field from the literature \cite{soar}, \cite{hardouin2023}, \cite{sweep_your_map}, \cite{racer}, \cite{fast_forests}.

The experimental results presented in this paper took place in the following simulated worlds:
\begin{itemize}
    \item House: a 45x40x20m outdoor open field urban environment composed of a house, two cars, and a complex playground structure. All elements are dispersed within the map and disconnected, except for the ground surface. Object shapes are quite complex. 
    \item Structure: a 26x26x15m multi-level structured urban terrain composed of a combination of boxes and stepped platforms (see Fig.~\ref{fig:qualitative_mesh}).
\end{itemize}

\begin{table}[tb]
\centering
\begin{tabular}{|l|c|}
\hline
\textbf{Parameter} & \textbf{Value}   \\
\hline
Robot Radius           & 1.0 m   \\
Voxel Size $d_s$             & 0.25 m \\
Lidar Min/Max Range $r_{min}, r_{max}$        & 0.42m / 10m   \\
Vertical Min/Max Angle     & -35° / 30°  \\
Maximum velocity $vel_{max}$             & 4.5 m/s       \\ 
Maximum absolute acceleration $acc_{max}$     & 4.8 m/s²   \\  
\hline
\end{tabular}
\caption{Simulation parameters 
\vspace{-0.2cm}}
\label{tab:simulation_parameters}
\end{table}

\begin{table*}[t]
\centering
\footnotesize
\setlength{\tabcolsep}{3pt}
\renewcommand{\arraystretch}{0.95}
\begin{tabular}{l|ccccc|ccccc}
\hline
& \multicolumn{5}{c|}{\textbf{House}} 
& \multicolumn{5}{c}{\textbf{Structure}} \\
\hline
Method 
& Makespan & Av. Path & Cov. (\%) & Prec. (\%) & F-Score
& Makespan & Av. Path & Cov. (\%) & Prec. (\%) & F-Score \\
\hline

Ours ($d^* = 4.5m$)
& \textbf{258.2} & \textbf{200.9} & \textbf{80.73} & \textbf{69.06} & \textbf{74.44}
& \textbf{289.4} & \textbf{236.6 }& \textbf{95.48 }& \textbf{87.03} & \textbf{91.06 }\\

Ours w/o imbalance corr. ($d^* = 4.5m$)
& 329 & 260 & \textbf{81.6} & \textbf{69.13} & \textbf{74.84}
& 336 & 279 & \textbf{96.03} & \textbf{87.01} & \textbf{91.29} \\

FAME \cite{fast_forests}
& 457 & 274 & 43.63 & 37.25 & 40.18
& 586 & 371 & 78.23 & 70.69 & 74.26 \\

RACER \cite{racer}
& 430 & 262 & 41.66 & 37.35 & 39.38
& 544 & 434 & 74.94 & 73.83 & 74.38 \\

\hline
\end{tabular}
\caption{Quantitative results in the House and Structure environments with 3 UAVs.}\label{tab:quantitative_results}
\end{table*}

    \begin{figure}
        \centering
        \includegraphics[width=\linewidth]{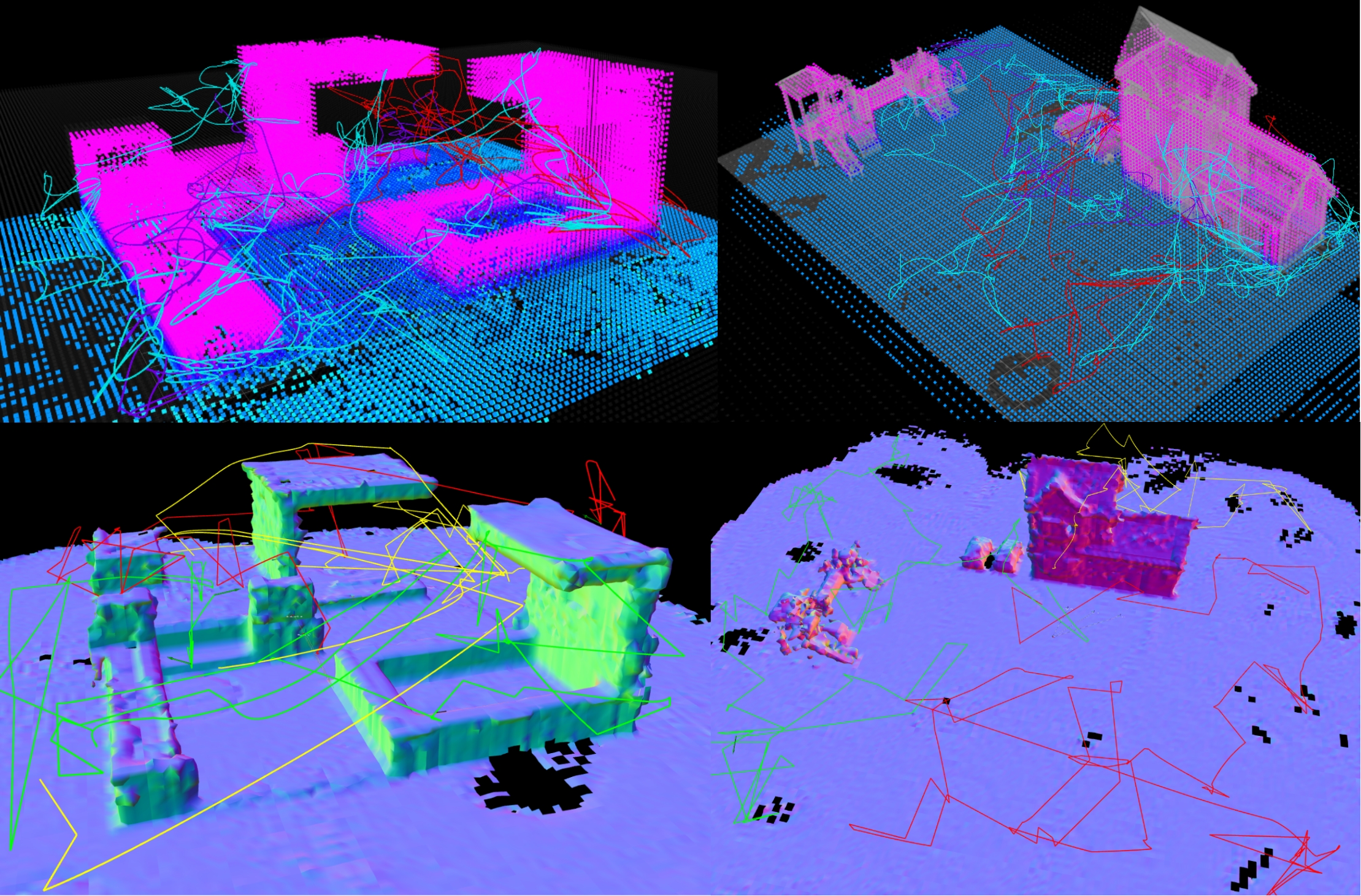}
        \caption{Comparison of the paths with a team of 3 robots for FAME (up) and ours (down) in both environments. In FAME images, the built map is in pink and the ground truth in grey. }
        \label{fig:qualitative_path}
    \end{figure}

We selected these environments for their challenging and complementary characteristics that can easily generate back-and-forth motions. In "House", the scattered objects of interest, and in "Structure", the multiple levels with hidden ceiling surfaces, narrow gaps, and concave corners create complex topologies where the robot must efficiently prioritize the views to select. 
For both environments, we imposed a mission time limit of 15 minutes.

\begin{figure}
        \centering
        \includegraphics[width=0.73\linewidth]{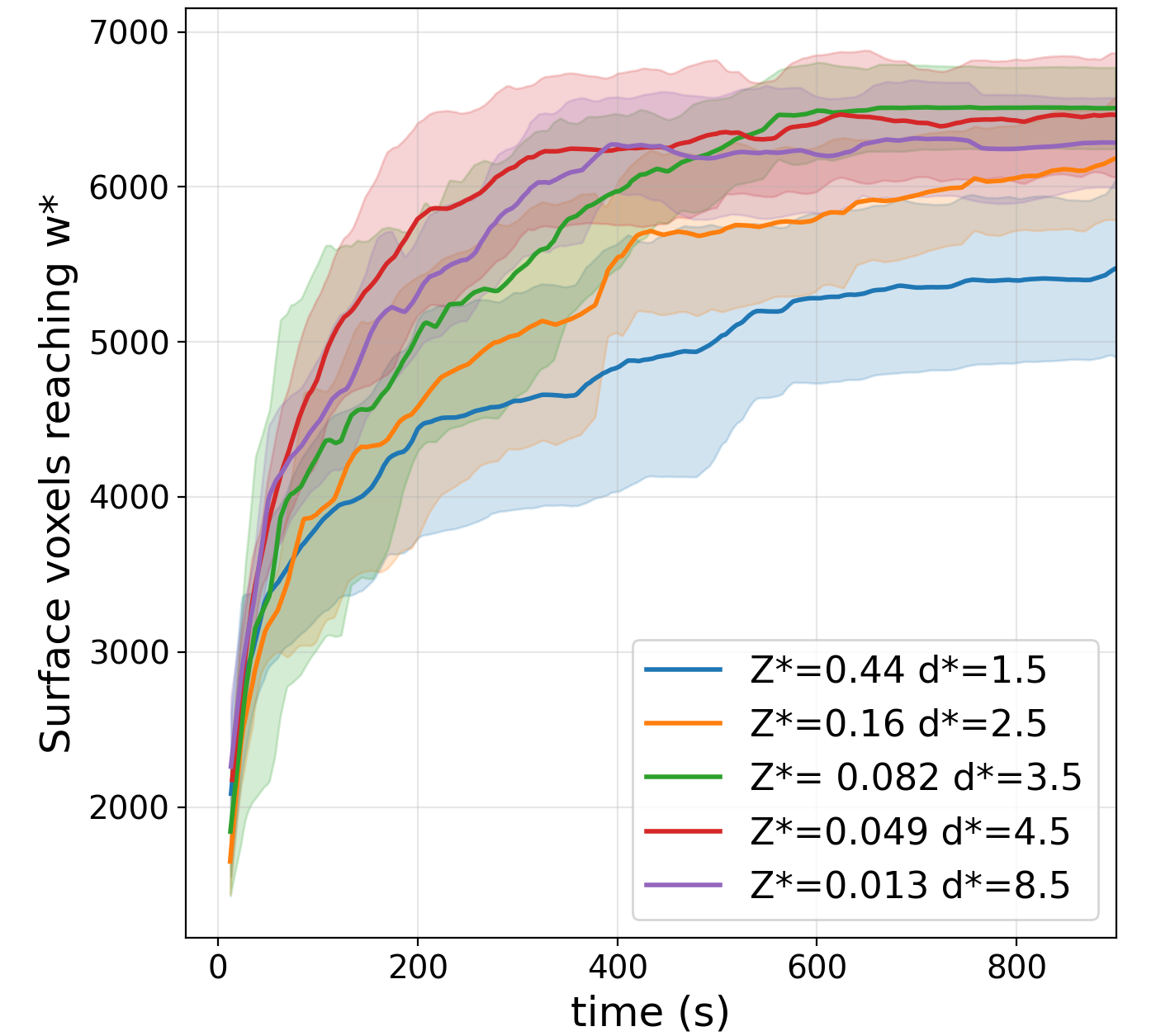}
        \caption{Number of voxels reaching the confidence objective $w^*$ with corresponding values of $d^*$ for a team of 3 robots in the Structure environment.}
        \label{fig:simulation_comp}
    \end{figure}

All of our experiments are run with $N=3$ UAVs. Across all scenes, we keep the same values of $\alpha = \beta = 1$ and $\gamma = 2$. The spatial reconfiguration of the fleet computed within the global centroid-based replanning scheme (detailed in \ref{sec:approach}-C) is only valid until the next map update, which may alter the active surface element set and shift the centroids. For this reason, we want the UAVs to move toward their assigned subregion immediately after receiving the solution, but not to be unnecessarily limited to this portion of the map for several subsequent steps. Thus, we enforce the corrected selection based on the centroids only for $k = 2$ steps, which we found to be the most efficient value for enabling fast reconfiguration without negatively impacting exploration speed. 
We report ours with $d^* = 4.5m$, which corresponds to half the maximum range of the sensor $r_{max}$, and showcased the best trade-off between speed and accuracy in our experiments (see Fig.\ref{fig:simulation_comp}). Other simulation parameters are detailed in Table \ref{tab:simulation_parameters} and we use these values in all methods. 

\subsection{Metrics}
The performance of the approaches is evaluated on different criteria. First, we use the average distance traveled and particularly the makespan of path lengths as a proxy for mission time. Secondly, we extract the meshes from the maps at the end of the mission, using the Marching Cubes algorithm, and compare them to the ground-truth meshes of the simulated environments. To this end, we implemented the metrics defined in the 3D reconstruction benchmark \cite{benchmark_3d}: recall (or coverage), precision, and F-score, measuring completeness, accuracy, and a combination of both, respectively. We consider a point of the original mesh to be covered if its closest point from the obtained map is at a distance of less than a voxel diagonal $\sqrt{3/2} \ d_s$. 
Since the methods are not deterministic, all reported values are averages over 10 runs. 

\subsection{Benchmark and Comparison Analysis}

The first conclusion that can be drawn from the set of comparative results presented in Table~\ref{tab:quantitative_results} is that our proposed approach outperforms the other two methods in all map fidelity metrics for both environments while maintaining significantly smaller path lengths. 
In "House", in our reconstruction-oriented setting, RACER and FAME remain below 45\% of total coverage while we cover more than 80\%, i.e., an 82\% increase with a makespan 40\% smaller. In "Structure", the gap is less pronounced but still significant: despite a makespan at least 47\% shorter, our coverage and precision are respectively 21\% and 22\% higher. However, the averaged results of RACER and FAME hide a greater variability than ours, which can be attributed to the similar view generation strategy that both methods share. When generating a new viewpoint, they sample randomly within a cylinder around the target frontiers, which can lead to very different paths each run. This method can also produce poor observation angles for surface observation. Because RACER and FAME are exploration approaches that also tend to focus on volume coverage more than the surface when planning paths and determining views, holes in the final mesh and sparse surfaces were to be expected. This difference in terms of mesh quality can, for instance, be observed in the illustrative example shown in Fig.~\ref{fig:qualitative_mesh}. However, it is interesting to note that we obtain better results also in terms of path efficiency and exploration speed.
An analysis of RACER's and FAME's paths shows a difficulty in dispersing efficiently in space with significant overlap and redundancy, which leads to longer paths and repeated traversal (see Fig.~\ref{fig:qualitative_path}). In contrast, our method allows dividing the environment online, resulting in improved coordination. Visible in both environments, this is especially noticeable in "House", where the overlap between paths is minimal. 
Additionally, we evaluated the impact of $Z^*$ on reconstruction speed: we report in Fig.~\ref{fig:simulation_comp} the number of voxels reaching the aimed confidence value $w^*$ over time. Different values were selected to force $d^*$ at the upper, lower, and in between the limits of our depth sensor range. As expected, a low $w^*$ value leads to aggressive exploration and a high value to a more careful one. We can notice that, while reducing $Z^*$ improves exploration speed, this gain plateaus when $Z^*$ becomes very small. This effect can be attributed to both the structure of the environment and the size of the robot team. 
A similar analysis for a single UAV can also be found in our previous work \cite{Sportich2026ICUAS}.

\subsection{Ablation Studies}

To validate the relevance of the different modules of our system, we studied their impact on the performance in the reconstruction task. We ran experiments with and without the imbalance correction module to assess its efficiency, and the results are reported in 
Table~\ref{tab:quantitative_results}. The two solutions provided an almost identical coverage and accuracy, but the method implementing the balance correction showed an average path length 23\% smaller in the House environment and 15\% in the Structure environment. Regarding the makespan, the gap is also significant: 21\% on House and 13\% on Structure. 
We noticed that the imbalance correction is activated mostly in two cases: when UAVs are close and thus are competing for local active surface elements, and when a UAV finishes exploring a subregion of the map and has no active surface elements in the vicinity. The former occurs essentially at launch, while the latter toward the end of the mission.

\section{Conclusion}
This paper presented a quality-adaptive 3D reconstruction planning framework for cooperative UAV teams operating in unknown environments. By explicitly incorporating a TSDF-based confidence measure into a multi-robot next-best-view planning, the proposed approach allows the online generation of viewpoints that are consistent with user-defined reconstruction fidelity requirements, while maintaining efficient path execution. In order to further improve the coordination among the fleet, we presented an additional clustering-based area repartition mechanism that is sparsely triggered by unbalanced UAVs' configurations with respect to highly informative regions. Resulting in a global workload redistribution, the proposed solution allows the fleet to better adapt to evolving map structures.
To evaluate the performance of our approach, we provided comparative results in complex simulated environments. These results demonstrated a higher reconstruction quality and shorter paths compared to state-of-the-art multi-UAV exploration methods, while reducing redundant observations. Furthermore, an ablation study proved the significant improvement produced by the sparse redistribution of areas to explore.

In the future, we intend to adapt our quality criterion to different 3D representations, such as Gaussian Splatting or NeRF. We also plan to evaluate the impact of dynamic adaptive quality requirements during reconstruction planning, for instance, when triggered by a semantic mapping module.

\bibliographystyle{IEEEtran}
\bibliography{biblio}

\end{document}